\documentstyle[fi12pt]{article}

\newcommand{\Bem}[1]{}
\Bem{title
Reasoning and Facts Explanation in Valuation Based Systems
}
\Bem{shorttitle
Reasoning and Facts Explanation in VBS
}

\Bem{author
S.T. Wierzcho\'{n}, M.A. K{\l}opotek,  M. Michalewicz
}
\Bem{address
Institute of Computer Science\\ 
Polish Academy of Sciences\\ 
01-237 Warszawa, ul Ordona 21\\
e-mail:   stw{@}ipipan.waw.pl,      klopotek{@}ipipan.waw.pl,\\
        michalew{@}ipipan.waw.pl\\
Poland}
\date{}

\begin{document}
\begin{center}%
{\Large\bf Reasoning and Facts Explanation in Valuation Based Systems
} \bigskip

\begin{tabular}{c}%
S.T. Wierzcho\'{n}, M.A. K{\l}opotek,  M. Michalewicz\\
Institute of Computer Science\\ 
Polish Academy of Sciences\\ 
 Warszawa, Poland\\
e-mail:   stw{@}ipipan.waw.pl,      klopotek{@}ipipan.waw.pl,\\
        michalew{@}ipipan.waw.pl\\
\hspace*{0.4\hsize}   \\
 \end{tabular}

\setcounter{footnote}{0}
\end{center}

\begin{Abstract} 
In the literature, the 
  optimization problem to identify a set of
composite hypotheses H, which will yield the $k$  largest $P(H|S_e)$ where a
composite hypothesis is an instantiation of all the nodes in the network
except the evidence nodes \cite{KSy:93} is of significant interest. 
This problem is called "finding
the $k$ Most Plausible Explanation (MPE) of a given evidence $S_e$ in a
Bayesian belief network".          
The problem of finding $k$ most probable hypotheses is 
 generally NP-hard
\cite{Cooper:90}. Therefore in the past various simplifications of the task by
restricting $k$ (to 1 or 2), restricting the structure (e.g. to singly
connected networks), or  shifting the complexity to spatial domain have been
investigated. 

A genetic algorithm  is proposed in this paper to overcome some of these
restrictions while stepping out from probabilistic
domain onto the general Valuation based System (VBS) framework is also
 proposed by generalizing  the genetic algorithm approach to the
realm of Dempster-Shafer belief calculus.

\end{Abstract}

\normalsize
\section{Introduction}

        Bayesian network, BN for brevity, is a powerful model for
probabilistic reasoning \cite{Pearl:88}. From a formal standpoint
the BN is a triplet BN = (X*, E*, P*) where G = (X*,E*) is a
directed acyclic graph, and P* is a set of conditional
probabilities p(x|Pa(x)) where Pa(x) stands for the set of
parents of x $\in $ X* in the graph G. The graph G represents
$qualitative$ interrelationships among the variables specified
in the set X*, while P* gives a $quantitative$ description of
these interrelationships. Knowing the conditionals P* we can
express the joint probability distribution,
$p(X*)=p(x_1$,$x_2$,...,$x_n$$)$, as the product of all the
probabilities $p(x|Pa(x))$. Note that the symbol $p(x|Pa(x))$ is
a shorthand of the next formula: $p(x=a|x_i=b_i,...x_k=b_k)$,
where {$x_i,...,x_k$} = $Pa(x)$ and defined for all the values
$a,b_i,...,b_k$ ranging over the {\it discrete } domains of the
corresponding variables. 

        Now, the problem of the probabilistic reasoning can be stated
as follows. Assume that E $\subset $ X* is a set of $clamped$
variables, i.e. the variables with known values. Our task is to
find the conditional probability

\begin{equation} \label{new1}
        p*(x|E) = p(X*)/p(E)                                         
\end{equation}

for a variable x $\in $ (X* - E). 

        On the other hand, knowing the values of the variables
specified in the set E we may be interested in the values of the
remaining variables that provide the maximal value of the joint
probability distribution. That is, if E = {$x_1,x_2,...,x_k$}
and $x_i = a_i  ^{*}$ , i = 1..k, we are searching for such values
$x_j = b_j ^{*}$ , j = (k+1)..n, that

\begin{eqnarray} \label{new2}
        p(x_1=a_1 ^{*},...,x_k=a_k ^{*},x_{k+1}=b_{k+1} ^{*},...,x_n=b_n ^{
*})= \nonumber\\
        \max\limits_{x_{k+1},...,x_n}p((x_1=a_1 ^{*},...,x_k=a_k ^{*},x_{
k+1}=b_{k+1},...,x_n=b_n)                                      
\end{eqnarray}

The problem (\ref{new2}) is referred to as the facts explanation as the
values $x_j = b_j ^{*}$ , j = (k+1)..n together with $x_i =
a_i ^{*}$ , i = 1..k form the most probable configuration. More
precisely the equation (\ref{new2}) defines so-called first most probable
explanation, or $1-MPE$  for short. In quite similar way we can
define $k-MPE$ yielding k-th largest value of the $p*(x|E)$.

        It appears that both the problems (\ref{new1}) and 
(\ref{new2}) are extremely
difficult from the numerical standpoint. To be illustrative
assume that X* contains 56 binary variables. In this case the
joint probability distribution consists of $2 ^{56}$ values. If
our computer can calculate the terms of the probability
distribution for one million values per second, then it will
only take our computer 2283 years to come up with the whole
distribution function! 

        Fortunately both the problems can be solved in a reasonable
time and without computing the joint probability distribution.
The first solution, restricted to the case when G is a {\it tree }
was proposed by J. Pearl~\cite{Pearl:88}, and next it was extended to the case
when G is a graph by S.L. Lauritzen and D.J. Spiegelhalter~%
\cite{Lauritzen:88}.
Further improvement to this problem was proposed by F.V. Jensen.
Almost at the same time G. Shafer and P.P. Shenoy~\cite{Shafer:87} 
 have
considered similar problem of computing a belief function (i.e.
a generalization of a probability function) for a variable x
$\in $ X* given a set of clamped variables. Later results of A.
Kong led P.P. Shenoy to the notion of so-called Valuation Based
System, or VBS for brevity~\cite{Shenoy:92}. Within such a system
we can represent knowledge in different domains including probability
theory, Dempster-Shafer theory, possibility theory an so on.
More recent studies show that the framework of VBS is also
appropriate for representing and solving Bayesian decision
problems and optimization problems 
\cite{Shenoy:94}. The graphical representation of a problem is called a
{\it valuation network}, and the method for solving problems is
called the {\it fusion algorithm}~\cite{Pearl:86}.

        In this paper we briefly describe the idea of a VBS and we show
how it can be used in the reasoning process and in the fact
explanation problem  

        Up to this moment we have no satisfactory algorithms to verify
the "summarized expert's experience" against a detailed
knowledge (about a population) that is stored in a database.
Apart already mentioned papers, the practical attempt to solve
this problem was developed by our research group in the form of
the SEAD\_1 system reported in 
\cite{Michalewicz:91}. 
The system incorporates various tools supporting a
researcher in different parts of the cycle
observation-generalization-theory-exertion. The main functions
of the system are: storing observations and results of
experiments (data base), storing acquired theoretical
generalizations (knowledge base) and tools supporting knowledge
acquisition (statistical data analysis system, learning from
examples system, cf. 
\cite{Michalewicz:Michalewicz:91} and
knowledge verification (against new incoming experimental
results in the database or against new single cases). The main
feature of our solution is the implementation of all system
components into one organism who's heart is database pumping and
absorbing data. The next, just finished, version of the system
relies upon introducing different model of knowledge base. The
system SEAD\_2 is the experimental computer aided decision-making
system based on mixed, probabilistic/Dempster-Shafer, knowledge
representation, and it fulfills the next requirements:

\begin{itemize} 
\vspace{-2ex}\itemsep=-1ex  
\item representation of joint belief function as a belief network
(knowledge base),
\item 
 representation of joint belief function as a random sample
(data base),
\item 
 calculation of marginal belief function in relation to
observed facts (consultation session),
\item 
 automatic extraction of belief function from random sample
(knowledge generation)
\end{itemize}

        Although we implement quite effectively so-called message
passing algorithm \cite{Shenoy:90} the system was oriented towards
computing marginals for the nodes of the network only.
Frequently we are interested in finding joint probability
functions (or valuations if we use the language of VBS's) for
different subsets of the set of variables. This problem was
firstly solved by Xu \cite{Xu:95}, but it is possible to find more
general and more effective solution. In this
paper we present this general idea, and next we focus on the
problem of finding so-called most probable explanation of a
given set of observations.

\section{Valuation Based Systems}

Network-based approaches are at present the widely accepted
approaches for uncertainty processing and reasoning. Among them
Bayesian networks, \cite{Pearl:88}, designed for the case of
probabilistic reasoning, and more general valuation based
systems, or VBS for short, \cite{Shenoy:92}, are most popular.
From a graphical point of view Bayesian networks are directed
acyclic graphs, while VBS's are hypergraphs. Since any Bayesian
network can be translated to an equivalent VBS assume that we
start from a given Bayesian network which can be regarded as a
triplet (X*, E*, P*) where

X* = $\{x_1,x_2,...,x_m\}$ is a set of variables. With each
variable $x_i$ we associate its discrete frame $\Theta_i$. In
the sequel we shall write $\Theta$ to denote the Cartesian
product $\Theta _1\times \Theta _2\times ...\times \Theta _m$
and we shall write $\Theta (h)=\{\Theta _i|i\in h\}$. Thus
$\Theta =\Theta (X*)$.
\Bem{
\begin{figure}
\begin{center}
\input ki96bn.pic
\end{center}
\caption{A bayesian network representing the joint probability
distribution decomposition}
\label{bn}
 \end{figure}
}

E* is a set of directed edges over X* such that (X*, E*) is a
directed acyclic graph (DAG). Its nodes represent the variables
and the edges represent conditional dependency relationship
among variables,

P* is a set of conditional probabilities $p(X_i|Pa(X_i))$, i =
1..m, where $Pa(x_i)$ stands for the set of parents of i-th
variable in the DAG. When $x_i$ has no parents then $P(x_i)$
represents the unconditional probability of $x_i$. 
\Bem{Figure \ref{bn}
shows an example (taken from \cite{KSy:93}) Bayesian network.}

The joint probability of the set of variables X* is defined as

$$P(x_1,x_2,...,x_m)=\prod_{i=1}  ^mp(x_i|Pa(x_i))$$ 

The computation of full probability distribution is a time
consuming problem. For instance if we have 56 binary variables
then the total space of configurations, $\Theta $  consists of
$2 ^{56}$ elements. If our computer can calculate the probability
for one million configurations per second, then it will take
almost 2,283 years to come up with the whole joint probability
function. 

To overcome this difficulty we transform the Bayesian network
into a Valuation Based System which can be represented by the
triplet (X*, H*, V*) where H* is a family of subsets of the set
of variables X* and V* is a set of valuations.
\Bem{
 A valuation
network corresponding to the Bayesian network from Figure \ref{bn} is
given in Figure \ref{hg}.}
 In general valuations are primitives in the VBS
framework and, as such, require no definition. Intuitively they
represent some knowledge about the variables included in the
sets $h_i$ in H*, called {\it local universes}. Under probabilistic
case the local universes are defined as: $h_i=x_i\cup Pa(x_i)$.
With such a universe, $h_i$ , we associate the valuation
$v_i=p(x_i|Pa(x_i))$. To perform inferences with the set of
valuations we define two operations: combination ($\otimes $)
that corresponds to the aggregation of knowledge and
marginalization ($\downarrow $) that corresponds to the
coarsening of the knowledge. The join valuation $v$ for the set
X* is computed due to the equation 

\Bem{
\begin{figure}
\begin{center}
\input ki96hg.pic
\end{center}
\caption{A hypergraph representing the joint belief distribution decomposition
from the example}
\label{hg}
\end{figure}
}

$$v=\bigotimes_{i=1} ^mv_i$$ 
while the marginal valuation for some subset $h$ of X* is
computed from $v$ according to the equation

$$v(h)=v ^{\downarrow h}$$ 

It is easy to observe that under probabilistic context $\otimes
$ is equivalent to the product operator while $\downarrow $ is
equivalent to the summation. In other words $(\otimes
,\downarrow )=(\cdot ,\sum )$. Thus if  $h=\{x_j,...,x_k\}$ then

$$v(h)=\sum_{x\in (X ^*-h)}\prod_{i=1} ^m
v_i=(\bigotimes_{i=1} ^mv_i) ^{\downarrow h}$$ 

In the Dempster-Shafer calculus, \cite{Shenoy:94}, instead of
the probability distribution we use so-called basic probability
function, $m$. If $h$ is a subset of X* then as $m$ we regard a
set-function $m:2 ^{\Theta (h)}\rightarrow [0,1]$ such that
\begin{itemize}
\vspace{-2ex}\itemsep=-1ex  
\item[(m1)] $m(A)\geq 0$ for all $A\in 2 ^{\Theta (h)}$, 
\item[(m2)] $\sum \{m(A)|A\in 2 ^{\Theta (h)}\}=1$.
\end{itemize}

A belief function $Bel$ is defined as $Bel$:$2^{\Theta(h)} \rightarrow [0,1]$
so that $Bel(A) = \sum_{B \subseteq A} m(B)$.  
A plausibility function be $Pl$:$2^{\Theta(h)} \rightarrow [  0,1]$  with 
$\forall_{A \in 2^{\Theta(h)}} \  Pl(A) = 1-Bel(\Theta(h)-A )$/

Another very useful set function is so-called commonality
function, $Q$, that is computed from $m$ due to the equation
$Q(A)=\sum \{m(B)|B\supseteq A\}$. If $Q_1$ and $Q_2$ are two
commonality function, they are combined due to so-called
Dempster's Rule of Combination 

 $$(Q_1\otimes Q_2)(A)=k\cdot Q_1(A ^{\downarrow h_1})\cdot
Q_2(A ^{\downarrow h_2}),A\subseteq (\Theta (h_1)\cup \Theta
(h_2))$$ 
where $k ^{-1}=\sum \{(-1) ^{|A|+1}Q_1(A ^{\downarrow h_1})\cdot
Q_2(A ^{\downarrow h_2})|A\subseteq (\Theta (h_1)\cup \Theta
(h_2))\}$ is the normalizing constant such that $k\neq 0$. Here
e.g. $A ^{\downarrow h_1}$ stands for the projection of the set
$A$ to the set ${h_1}$ of variables. Similarly, if $Q$ is a
commonality function over the set of variables $h$ and $g$ is a
subset of $h$ then the marginalization is defined as follows 

$$Q ^{\downarrow g}(A)=\sum \{(-1) ^{|B-C|}Q(B)|B,C\subseteq \Theta
(h)$$ s.t. $C ^{\downarrow g}\supseteq A$, and $B\supseteq C\}$ 

Assuming that $\otimes $ is a commutative and associative
operation, and that $\downarrow $ is distributive with respect
to $\otimes $, consult  \cite{Shenoy:92}, we are able to define
so-called Message Passing Algorithm that is a universal tool for
making inferences under several uncertainty formalisms. To apply
this algorithm we must convert first the hypergraph (X*, H*)
into a secondary structure called Markov tree (i.e. an acyclic
hypergraph that covers the original hypergraph). Intuitively the
Message Passing Algorithm tells the nodes of a Markov tree in
what sequence to send their messages to propagate the local
information throughout the tree. The algorithm is defined by two
parts: $a fusion rule$, which describes how incoming messages
are combined to make marginal valuations and outgoing messages
for each node; and a $propagation algorithm$, which describes
how messages are passed from node to node so that all of the
local information is globally distributed. Just as propagation
takes place along the edges of the tree, fusion takes place
within the nodes. More formally, to compute the marginal $v(h)$
for a subset $h$ of X* (where $h$ is a member of H* or a subset
of some set from H*) we must combine the original valuation
$v_h$ assigned to this subset and the messages sent by all its
neighbors in the tree: 

$$v(h)=v_h\otimes (\otimes \{M_{g\rightarrow h}|g\in N(h)\})$$ 
where $N(h)$ stands for the set of neighbors of $h$ in the
tree, and $M_{g\rightarrow h}$ is the message sent from $g$ to
its neighbor $h$, which is computed by projecting on the set
$g\cap h$ the combination of $v_g$ and the messages sent by the
neighbors $g$ except $h$:

$$M_{g\rightarrow h}=(v_g\otimes (\otimes \{M_{k\rightarrow
g}|k\in (N(g)-h)\})) ^{\downarrow (g\cap h)}$$ 

This algorithm can be easily adopted to solving the 1-MPE
problem. Under probabilistic and also in Dempster-Shafer calculus 
(under assumptions that are listed    
in subsection 3.2.) 
 context we take the pair of
operators $(\otimes ,\downarrow )=(\cdot ,max )$. 
However, we are sometimes interested in flexibility of this solution, so would
like to know the subsequent 2nd,..,kth best solutions. 
But it is hard to extend this pathway of solution to the 
 general k-MPE problem. Therefore we decided to try out the genetic 
approach. 

\section{Genetic Algorithm for Finding MPE.}

 Following e.g. \cite{Michalewicz:94} we define general genetic
algorithm as a kind of a probabilistic algorithm which maintains
a population of individuals P(t) = $\{x_1 ^t,x_2 ^t,...,x_n  ^t\}$
for iteration t. Each individual represents a potential solution
to the problem at hand, and, in any genetic program, is
implemented as some data structure S. Each solution
x$_{\mbox{i}}$(t) is evaluated to give some measure of
"fitness". Then, a new population - iteration (t+1) - is formed
by selecting the more fit individuals. Some members of the new
population are recombined, i.e., transformed by means of two
"genetic" operators to a new form. These operators are: (1) the
unary mutation operator that create new individuals by a small
changes in a single individual, and (2) the n-ary, where n $\geq
$ 2, crossover operator, which creates new individual by
combining parts on the n individuals. After some number of
iterations the program converges, and the best individual
represents the optimum solution.The general implementation of
this idea can be summarized in the form of the next pseudocode
(see \cite{Michalewicz:94} for details):

 \begin{verbatim} 
Procedure GeneticAlgorithm 
begin t:=0; 
   Initialise P(t); 
   Evaluate P(t); 
   while (not termination-condition) do 
   begin 
      t:=t+1; 
      Select P(t) from P(t-1); 
      Recombine P(t); 
      Evaluate P(t); 
   end; 
end; 

\end{verbatim}

This idea quite interestingly translates into a program for
finding most probable explanations both in probabilistic and
generalized, Dempster-Shafer, case. Below we present details
concerning the implementation.

\subsection{Probabilistic Case}

Here the situation is quite simple. The aim is to find such a
configuration $\theta  ^{*}=\{\theta _1 ^{*},\theta
_2 ^{*},...,\theta _{*}\}$  that 

$$p(x_1=\theta _1 ^{*},x_2=\theta _2 ^{*},...,x_m=\theta
_m ^{*}\})  =  \max _{\theta \in \Theta
}\prod_{i=1} ^kp(x_i=\theta _i|Pa(x_i)=\theta (Pa(x_i))$$ 

\noindent
where $\theta (Pa(x_i)$ stands for the projection of the
configuration $\theta $ onto the set of variables $Pa(x_i)$.
Hence as an individual we take simply a vector x = $\theta $,
where each member $\theta_i $ of $\theta $ takes its values from
the domain of the variable $x_i $. Hence we depart from the
"standard" binary representation of the individuals. This
however fastens computations. 

The mutation and crossover operators are implemented in almost
standard way. There is a nuance, however. Frequently we ask for
an MPE already knowing values of some variables. If C stands for
the set of clamped variables, then, after the recombination
phase the values at positions corresponding to the variables
form the C must remain unchanged. There is a number of
strategies to attain this goal. But in our implementation we
used the next one. In case of mutation denote $p_m$ the
probability of mutation. Then for each element of the individual
x and such that it does not correspond to a variable from the C
we generate a random real number $r$ from the unit interval. If
$r < p_m$ we replace this element by another from the domain of
the corresponding variable. The crossover is obvious also, that
is we choose two parent individuals and the crossing point, and
finally we replace the parents by a pair of their offsprings. 

 The fitness of each individual is computed by means of the
maximized function $p(\cdot )$. The values of a conditional
probability $P(x_i|Pa(x_i))$ are stored in the structure
{\it universe }  defined below\\

\noindent
\verb|universe = record| \\  
\verb|variables: SetOfVariables;| (*i.e. $variables = \{X_i\cup Pa(X_i)\}$*)\\
\verb| card: byte;| (*cardinality of the set $variables$ *) \\
\verb|valuations: array [1..k] of real;| \\
\verb|end;| \\

In our implementation we assume that the variables are numbered
consecutively from 1 to $variables$ (number of variables). Further
the variables stored in the field $variables$ are written in the
next order: first is given he conditioned variable and next the
conditioning variables.

Assuming that the domains of all the variables are stored in the
table referred to as {\it domains } and that individuals are
represented as arrays called $config$ the value of the
conditional probability $P(x_i|Pa(x_i))$ for a given individual
x are computed by means of the next function 

 \begin{verbatim}

Function FindValuation(x: config; sp: universe): real; 
var j,position,size: integer; 
begin  
   with sp do  
   begin    
      position := x[variables[card]];    
      size := domains[variables[card]];     
      for j:= card-1 downto 1 do    
      begin      
         position := position + size*(x[variables[j]]-1);      
         size := size*domains[variables[j]];    
      end;    
      FindValuation := valuations[position];  
   end; 
end; {FindValuation} 
\end{verbatim}

The set of all universes is stored in the array referred to as
$universes$. Now it is easy to write down the function that
computes fitness of an individual x. Besides the already
mentioned function FindValuation we need a boolean function
Blocked. It is needed when we compute k-MPE. That is a solution
to an i-MPE problem, $i < k$, is removed on a stack of blocked
configurations. Now, looking for the k-th explanation we check
is the individual x belongs to this stack. If so, we mark it as
blocked and we search for another individual. Below we present
the pseudocode of the function.

 \begin{verbatim}

Function ObjectiveFunction(x: config; pro: cuniverses): real; 
var j: integer; temp: real; 
begin 
  if Blocked(x) then ObjectiveFunction := 0 else 
  begin 
    temp := 1; 
    for j := 1 to nvaluations do 
      temp := temp*FindValuation(x,pro[j]); 
    ObjectiveFunction := temp; 
  end; 
end; {ObjectiveFunction}

\end{verbatim}

\subsection{Case of Dempster-Shafer Calculus}

Extension of this approach to Dempster-Shafer theory of evidence is not
straightforward. First of all we do not have the general concept of Bayesian
network
decomposition of the joint belief distribution but rather a decomposition in
terms of a hypergraph \cite{Shenoy:90}. Though a Bayesian network-like
decomposition has been proposed by Cano et al \cite{Cano:93}, but most belief
functions cannot be decomposed that way.  

Contrary to probabilistic case, valuations are not defined exclusively for
elements of $\Theta$, but rather  for  elements  of  the 
power set of $\Theta$.

 Further, the observed variables  do not need to be clamped to a
single value, but they can represent a set of values.

Then we have the
difficult choice which configuration evaluation function to optimize: mass
function, belief function, plausibility function and the commonality
function. They represent various aspects of partial ignorance of 
reasoner's knowledge and it is hard to  discard  any  of  them  a 
priori.

If $f(x)=(\bigoplus_i Bel_i)(x)$ should be our target function, then in
general f(x) is not a function of $Bel_1 (x)$, $Bel_2 (x)$, .... because of
Dempster rule of evidence combination, where evidence from supersets 
contributes to subsets of values upon combination of evidence.
 Last not least we have a difficult problem of how to understand a
configuration: as a cross product of singleton values of all the variables, as
a  cross product of subsets of domains of all variables, or as a subset of the
cross product of all the domains of all variables.

We had to make some choices and we ca only justify them, not to prove their
validity in mathematical sense.
\begin{itemize}
\vspace{-2ex}\itemsep=-1ex  
\item 
First we decided that we want to find $k$ most optimal singleton
configurations.
We just understand that the belief function represents an imprecise 
information on some singleton configurations. 
\item 
We assumed that we want to find most plausible configurations that is ones
that cannot be rejected. Just we find an explanation good if it is hard to
prove it wrong rather then being easy to be proved right. 
\end{itemize}
These two choices greatly simplified our task.  One can   prove that then
the following target function gets its maximum at  the configuration 
solving the $k$-MPE problem: 
$$f(\underline{x})=(\bigoplus_{i=1}^{n} Q_i)(\underline{x})= 
\prod_{i=1}^{n}Q_i(\underline{x} ^{\downarrow Space(Q_i)})$$
($Space(Q_i)$ - the set of variables for which $Q_i$ is defined)
subject eventually to constraints that 

\begin{itemize}
\vspace{-2ex}\itemsep=-1ex  
\item[(1)] the set of observed variables E has
values restricted to the observed  sets of values 
\item[(2)] when we are looking for
ith most probable solution,
 we punish f() so that f(\underline{x})=0 whenever \underline{x} is one of
already found 1st,2nd, (i-1)st most probable configurations configurations.
\end{itemize}
The configurations are represented and evaluated essentially in the same
way as in probabilistic case, because they were designed for hypergraph
representation of joint probability distribution. The genetic algorithm
procedures are also the same except that
mutation and cross-over operations are designed in such a way as to keep
values of observed ("clamped") variables within the observed value sets
(rather than restricting them to a single value). That is, compared
to probabilistic case, mutation is not forbidden at clamped variables but
rather the scope of mutation is restricted to the subset of the values of the
given variable. 

\begin{table}
\caption{$m$-functions of components in the decomposition of $Bel$}
\label{tm}
\begin{center}
\begin{tabular}[t]{|c|c|}
\hline
$m_1$ in vars    &      {A,B}\\ 
\hline
\{(b1,a1)\}      &  0.20\\
\{(b2,a1),(b1,a2),\}      &  0.10\\
\{(b1,a1),
  (b2,a3),
  (b1,a2),
  (b2,a2)\}      &  0.45\\
\{(b1,a1),
  (b2,a3),
  (b1,a3)\}      &  0.25\\
\hline
$m_2$ in var &  {A} \\
\hline
\{a1\}     &  0.40\\
\{a2\}     &  0.05\\
\{a3\}     &  0.15\\
\{a1,a2\}     &  0.35\\
\{a1,a3\}     &  0.05\\
\hline
$m_3$ in vars   &       {C,B}\\
\hline
\{(c1,b1)\}      & 0.10  \\
\{(c1,b2),(c1,b1)\}      & 0.10 \\
\{(c1,b2),       
  (c2,b1)\}      & 0.70 \\
\{(c1,b1),
  (c1,b2),
  (c2,b2)\}      & 0.10  \\
\hline
$m_4$ in vars & {F,B} \\
\hline
\{(f1,b1)\}      & 0.05\\
\{(f2,b1)\}      & 0.10\\
\{(f2,b2)\}      & 0.32\\
\{(f1,b2)\}      & 0.03\\
\{(f2,b1),
  (f2,b2),   
  (f1,b2)\}      & 0.15\\
\{(f1,b1)
  (f2,b1)
  (f2,b2)\}      & 0.35\\
\hline
\end{tabular}
\end{center}
\end{table}
\begin{table}
\begin{center}
\begin{tabular}[t]{|c|c|}
\hline
$m_{7}$ in vars &  {G,F} \\
\hline
\{(g1,f1),(g2,f1),(g1,f2)\}      & 0.40\\%
\{(g1,f2)\}      & 0.08\\%
\{(g2,f1),(g1,f2)\}      & 0.20\\%
\{(g2,f2)\}      & 0.32\\%
\hline
$m_{8}$ in vars &  {J} \\
\hline
\{j1\}         & 0.46\\%
\{j2\}         & 0.54\\%
\hline
$m_{9}$ in vars &   {D} \\
\hline
\{d1\}         & 0.40\\%
\{d2\}         & 0.60\\%
\hline
$m_{10}$ in vars &  {H,G} \\
\hline
\{(h1,g1)\}      & 0.10\\%
\{(h1,g2)\}        & 0.18\\%
\{(h2,g1)\}         & 0.10\\%
\{(h2,g2)\}         & 0.02\\%
\{(h1,g1).  
  (h1,g2),
  (h2,g1),       
  (h2,g2)\}         & 0.20\\%
\{(h1,g1),
  (h1,g2)\}        & 0.40\\%
\hline
\end{tabular}       
\end{center}
\end{table}
\begin{table}
\begin{center}

\begin{tabular}[t]{|c|c|}
\hline
$m_5$ in vars & {E,C,D}\\
\hline
\{(e1,c1,d1)\}   & 0.20\\%
\{(e1,c2,d1),
  (e2,c2,d2)\}   & 0.05\\%
\{(e2,c1,d1)\}   & 0.10\\%
\{(e1,c1,d1),
  (e1,c1,d2),
  (e1,c2,d1),
  (e2,c1,d1),
  (e2,c1,d2),
  (e2,c2,d2)\}   & 0.05\\%
\{(e1,c2,d1), 
  (e1,c2,d2), 
  (e2,c1,d2)\}   & 0.30\\%
\{(e1,c1,d1),
  (e1,c1,d2),
  (e1,c2,d1),
  (e1,c2,d2),    & \\
  (e2,c1,d1),
  (e2,c1,d2),
  (e2,c2,d1),
  (e2,c2,d2)\}   & 0.30\\%
\hline
$m_{6}$ in vars & {I,F,J} \\
\hline
\{(i1,f2,j1)\}   & 0.10\\%
\{(i1,f2,j2)\}   & 0.10\\%
\{(i1,f1,j1),
  (i1,f2,j1)\}   & 0.20\\%
\{(i1,f1,j1), 
  (i1,f2,j1),
  (i1,f2,j2),
  (i2,f1,j2)\}   & 0.20\\%
\{(i1,f1,j1),
  (i1,f2,j1),
  (i1,f2,j2),
  (i2,f1,j2),
  (i2,f1,j1),
  (i1,f1,j2)\}   & 0.10\\%
\{(i1,f1,j1),
  (i1,f1,j2),
  (i1,f2,j1),
  (i1,f2,j2),
  (i2,f1,j2),
  (i2,f2,j2)\}   & 0.20\\%
\{(i1,f1,j1),
  (i1,f1,j2),
  (i1,f2,j1),
  (i1,f2,j2),   & \\
  (i2,f1,j1),
  (i2,f1,j2),
  (i2,f2,j1),
  (i2,f2,j2)\}   & 0.10\\%
\hline
\end{tabular}
\end{center}
\end{table}

Notice, however the difference that in probabilistic case f() was in fact
the sought ith most probable configuration probability itself, but in DST
case f() is only proportional to the  sought ith most plausible
configuration plausibility. 

If we would like to find the most optimal 
configuration not restricting ourselves to the singleton set, we would fall
into several triviality traps (not to mention explosion of complexity).
\begin{itemize}
\vspace{-2ex}\itemsep=-1ex  
\item
 If maximization of
plausibility function would be the criterion of
optimality, then always the universe set $\Theta$ would win because
$Pl(\Theta)=1$
and this is the maximum value of Pl-function. 
\item 
The same effect is achieved if
belief function is maximized   because $Bel(\Theta)=1$
and this is the maximum value of Bel-function. 
\item 
If we maximize the commonality
function then for every non-singleton configuration $C$, any configuration
$C'\subset C$ is at least as optimal as $C$.  
\item
Finding an optimal configuration
for the mass function would be a complex task because of the Dempster-rule of
evidence combination.
\end{itemize}

EXAMPLE: Let us consider the joint belief distribution $Bel$
(derived from the probabilistic example of Sy\cite{KSy:93})
 decomposed in
belief functions $Bel_i$, i=1,...,10 such that 
$$Bel = \bigoplus_{i=1}^{10}Bel_i$$
where the $m$ functions are given in Table \ref{tm}. 
\Bem{This decomposition can be depicted in terms of a hypergraph as in 
 Figure \ref{hg}.} 

To find the most plausible configuration explaining the observation that 
H equals h1, F equals f2, G equals g2, J equals  j2 and  A is either a1 or
a3, we first have to identify  the  Q  functions  values  of  all 
singleton sets
appearing in belief functions $Bel_i$, i=1,..., 10. The resulting
$Q$-functions are listed in Table \ref{tQ}.

\begin{table}
\caption{$Q$-functions of components in the decomposition of $Bel$ - 
listed only for singleton subsets of respective domains.}
\label{tQ}
\begin{center}
\begin{tabular}[t]{|c|c|}
\hline
$Q_1$ in vars    &      {A,B}\\ 
\hline
\{(b1,a1)\}      &  0.90\\
\{(b1,a2)\}      &  0.55\\
\{(b1,a3)\}      &  0.25\\
\{(b2,a1)\}      &  0.10\\
\{(b2,a2)\}      &  0.45\\
\{(b2,a3)\}      &  0.70\\
\hline
$Q_2$ in var &  {A} \\
\hline
\{a1\}     &  0.80\\
\{a2\}     &  0.40\\
\{a3\}     &  0.20\\
\hline
$Q_3$ in vars   &       {C,B}\\
\hline
\{(c1,b1)\}      & 0.30  \\
\{(c1,b2)\}      & 0.90 \\
\{(c2,b1)\}      & 0.70 \\
\{(c2,b2)\}      & 0.10  \\
\hline
$Q_4$ in vars & {F,B} \\
\hline
\{(f1,b1)\}      & 0.40\\
\{(f1,b2)\}      & 0.18\\
\{(f2,b1)\}      & 0.60\\
\{(f2,b2)\}      & 0.82\\
\hline
\end{tabular}
\begin{tabular}[t]{|c|c|}
\hline
$Q_5$ in vars & {E,C,D}\\
\hline
\{(e1,c1,d1)\}   & 0.55\\%
\{(e1,c1,d2)\}   & 0.35\\%
\{(e1,c2,d1)\}   & 0.70\\%
\{(e1,c2,d2)\}   & 0.60\\%
\{(e2,c1,d1)\}   & 0.45\\%
\{(e2,c1,d2)\}   & 0.65\\%
\{(e2,c2,d1)\}   & 0.30\\%
\{(e2,c2,d2)\}   & 0.40\\%
\hline
$Q_{6}$ in vars & {I,F,J} \\
\hline
\{(i1,f1,j1)\}   & 0.80\\%
\{(i1,f1,j2)\}   & 0.40\\%
\{(i1,f2,j1)\}   & 0.90\\%
\{(i1,f2,j2)\}   & 0.70\\%
\{(i2,f1,j1)\}   & 0.20\\%
\{(i2,f1,j2)\}   & 0.60\\%
\{(i2,f2,j1)\}   & 0.10\\%
\{(i2,f2,j2)\}   & 0.30\\%
\hline
\end{tabular}
\begin{tabular}[t]{|c|c|}
\hline
$Q_{7}$ in vars &  {G,F} \\
\hline
\{(g1,f1)\}      & 0.40\\%
\{(g1,f2)\}      & 0.68\\%
\{(g2,f1)\}      & 0.60\\%
\{(g2,f2)\}      & 0.32\\%
\hline
$Q_{8}$ in vars &  {J} \\
\hline
\{j1\}         & 0.46\\%
\{j2\}         & 0.54\\%
\hline
$Q_{9}$ in vars &   {D} \\
\hline
\{d1\}         & 0.40\\%
\{d2\}         & 0.60\\%
\hline
$Q_{10}$ in vars &  {H,G} \\
\hline
\{(h1,g1)\}      & 0.70\\%
\{(h1,g2)\}        & 0.78\\%
\{(h2,g1)\}         & 0.30\\%
\{(h2,g2)\}         & 0.22\\%
\hline
\end{tabular}       
\end{center}
\end{table}

Then the genetic algorithm can be started, with e.g.
variable clamped: H to h1, F to f2, G to g2 and J to j2, A 
restricted to \{a1,a3\}.

Final solution gave 
ObjectiveFunction equal 0.01100484 was 
reached for the chromosome (1,1,2,2,1,1,2,1,2,2) that is  the most plausible
configuration is:\\
 a1,b1,c2,d2,e1,f1,g2,h1,i2,j2. 

\section{Concluding Remarks}

Our genetic approach 
to solving the problem of 
finding the $k$ Most Plausible Explanation (MPE) of a given evidence $S_e$ 
within the general framework of Valuation Based Systems (VBS)
consists generally in a special design of the target
maximized function as:
$$f(\underline{x})=\prod_{i=1}^{n}P(x_i|Pa(x_i))$$ 
in probabilistic case and 
$$f(\underline{x})=\prod_{i=1}^{n}Q_i(
\underline{x} ^{ \downarrow Space(Q_i)})$$
in case of DST, where $P(x_i|Pa(x_i))$ and 
$Q_i(\underline{x} ^{ \downarrow Space(Q_i)})$
are component valuation in a graphoidal decomposition of joint belief
distribution, 
subject eventually to constraints that (1) the set of observed variables E has
clamped values or value sets (2) when we are looking for ith most
probable/plausible solution,
 we punish f() so that f(\underline{x})=0 whenever \underline{x} is one of
already found 1st,2nd, (i-1)st most probable/plausible configurations. 
Mutation and cross-over operations are designed in such a way as to keep
clamping of observed variables intact. We tested this approach against
results presented in the literature e.g. for single connected networks of Sy
\cite{KSy:93}. 

The  essential  gain  from  using  genetic  algorithm   approach, 
compared to other
approaches, consists however in the possibility of finding $k$ most
probable/plausible
explanations not only for singly connected networks
~\cite{KSy:93,Henrion:90},
 or biparite graphs\cite{Wu:90}, or
hypertrees, but also for multiply connected bayesian networks and for general
hypergraph representations of joint belief distributions, both in
probabilistic and DST case.  The complex task of transforming bayesian network
representation or general hypergraph representation into a hypertree or
another singly connected network, advocated by some authors (e.g.
\cite{Pearl:88,KSy:93}), is not necessary.

Further research is needed to exploit the full potential of this technique. On
the one hand the performance of the algorithm for lager networks, 
with 100 and
more nodes needs to be tested. On the other hand an interesting question is
the possibility of modifying cross-over and/or mutation operations to take
advantage from (conditional) independence information contained in the
graphoidal structures of decompositions.

\end{document}